\documentclass[conference]{IEEEtran}
\IEEEoverridecommandlockouts
\usepackage{cite}
\usepackage{amsmath,amssymb,amsfonts}
\usepackage{algorithm,algorithmicx,algpseudocode}
\usepackage{graphicx}
\usepackage{textcomp}
\usepackage{xcolor}
\usepackage{caption, subcaption}
\usepackage{url}
\def\BibTeX{{\rm B\kern-.05em{\sc i\kern-.025em b}\kern-.08em
    T\kern-.1667em\lower.7ex\hbox{E}\kern-.125emX}}

\newtheorem{definition}{Definition}

\begin{document}

\title{Uncertain Time Series Classification With Shapelet Transform\\
\thanks{This work is funded by the french Ministry of Higher Education, Research and Innovation, the LabEx IMobS3 and the CNRS PEPS project TransiXplore. Thank to the UEA \& UCR Time Series Classification Repository which provides the datasets used for our experiments.}
}

\author{\IEEEauthorblockN{1\textsuperscript{st} Michael Franklin MBOUOPDA}
\IEEEauthorblockA{CNRS, ENSMSE, LIMOS, F-63000 \\
\textit{University of Clermont Auvergne}\\
Clermont-Ferrand, France \\
michael.mbouopda@uca.fr}
\and
\IEEEauthorblockN{2\textsuperscript{nd} Engelbert MEPHU NGUIFO}
\IEEEauthorblockA{CNRS, ENSMSE, LIMOS, F-63000 \\
\textit{University of Clermont Auvergne}\\
Clermont-Ferrand, France \\
engelbert.mephu\_nguifo@uca.fr}
}


\maketitle

\begin{abstract}
Time series classification is a task that aims at classifying chronological data. It is used in a diverse range of domains such as meteorology, medicine and physics. In the last decade, many algorithms have been built to perform this task with very appreciable accuracy. However, applications where time series have uncertainty has been under-explored. Using uncertainty propagation techniques, we propose a new uncertain dissimilarity measure based on Euclidean distance. We then propose the uncertain shapelet transform algorithm for the classification of uncertain time series. The large experiments we conducted on state of the art datasets show the effectiveness of our contribution. The source code of our contribution and the datasets we used are all available on a public repository.
\end{abstract}


\section{Introduction}
The last decade has been characterized by the availability of measurements in a large and variate set of domains such as meteorology, astronomy, medicine and object tracking. Generally, these measurements are represented as time series \cite{uts_sim.1931D}, that means a sequence of data ordered in time. Time series classification is used in many applications such as astronomy, land cover classification and human activity recognition.  Meanwhile, there has been an increase of the number of methods for time series classification \cite{Shifaz2020ts_chief,bagnall2017great}. However, to the best of our knowledge, these methods do not take data uncertainty into account. Any measurement is subject to uncertainty that can be due to the environment, the mean of measurement, privacy constraints and other factors. Furthermore, even if uncertainty can be reduced, it cannot be eliminated  \cite{err_analysis_jrtaylor}. In some applications, uncertainty cannot be neglected and has to be explicitly handled \cite{sarangi2010dust}.

Shapelet based methods are one of the best approaches that have been developed for time series classification. A shapelet is a subseries that is representative for a class of time series. These methods are especially appreciated for their interpretability, their robustness and their classification speed \cite{ye2009time}. A typical shapelet-based approach for classification can be summarized in three steps:
\begin{itemize}
	\item the first step is the extraction of shapelets. This step can be seen as a feature selection,
	\item the second step is the shapelet transformation. Here, the feature vector of each instance in the dataset is computed,
	\item the third and last step consists in training a supervised classifier using the generated feature vectors.
\end{itemize}

Almost every time series classification methods are built by coupling a similarity measure with a supervised classifier. We follow this pattern in this paper to build an uncertain time series classifier. We are not aware of any existing method in the literature for the classification of uncertain time series. 

Our contribution is as follows, we first propose an uncertain dissimilarity measure based on Euclidean distance. Secondly we use it to build the uncertain shapelet transform algorithm, which is the shapelet transform algorithm adapted to the classification of time series with available uncertainty information.

The rest of this paper is organized as follows: background and related works are presented in section \ref{sec:related_works}. In section \ref{sec:UED}, we present a new uncertain dissimilarity measure called UED, and in section \ref{sec:u_shapelet_classification}, we present the uncertain shapelet transform algorithm (UST). Section \ref{sec:experiments} is about experiments and section \ref{sec:conclusion} finally concludes this paper.

\section{Background and related works}\label{sec:related_works}
In this section we define the basic concepts and present the related works.

\subsection{Background}

\begin{definition}[Time series] A time series T of length $m$ is an ordered sequence of $m$ observations $t_i$.
	\[ T=<t_1, t_2,...,t_m> \]
\end{definition}

\begin{definition}[Uncertain time series] An uncertain time series (UTS) is a time series in which the observations are uncertain.
\end{definition}

Uncertain observations can be modeled with the multiset-based or the probability density function-based (PDF-based) modeling: 
\begin{itemize}
	\item In the \textbf{multiset-based} model, each $t_i$ is represented as a finite set of the possible values. By replacing each $t_i$ in $T$ by a single value from the set of its possible values, we obtain a realization of $T$.
	\item In the \textbf{PDF-based} model, $t_i = \hat{t_i} \pm \delta t_i$, where $\hat{t_i}$ is the best guess of the exact value and $\delta t_i$ is a positive number representing the maximum deviation from the best guess. In other words, $t_i$ is a random variable following a particular distribution and its exact value is somewhere in the interval $[\hat{t_i} - \delta t_i,\hat{t_i} + \delta t_i]$. 
\end{itemize}

\subsection{Related work on similarity measures for uncertain time series}
Given two uncertain time series, a probabilistic similarity measure computes the probability that the magnitude of the similarity between them is not greater than a user defined threshold \cite{uts_sim.1931D}. They are mostly used in the context of time series database querying. For multiset-based UTS, MUNICH \cite{munich2009} can be used. For pdf-based UTS, PROUD \cite{sarangi2010dust} can be used. 

Unlike probabilistic similarity measures, uncertain distances compute the dissimilarity between uncertain times series. When the distribution of the uncertainty is known DUST \cite{sarangi2010dust} can be used to compute the distance between UTS. Otherwise, the dissimilarity measure FOTS \cite{fots:siyou} can be used.

A common limitation of FOTS and DUST is that they define the distance between two uncertain time series as a real number, without any uncertainty information on that number. It is not possible to compare uncertain measures with $100\%$ reliability. For this reason, and because applying FOTS or DUST  is not always possible in practice, we propose in section \ref{sec:UED}, UED, an uncertain dissimilarity measure that makes no assumption on the distribution of the uncertainty and outputs the dissimilarity value with a confidence interval.

\subsection{Related works on time series classification with shapelets}
Basically, a shapelet is a subsequence that is common to time series from the same class label; it is called a representative subsequence for that class. Hence, given a new time series to be classify in a particular class label, the only thing to do is to check if the shapelet of that class label appears to be a subsequence of the time series. 

The first ever time series classification using shapelets is the shapelet decision trees \cite{ye2009time}. This algorithm constructs a decision tree in which attributes are shapelets. The value of each attribute for a given time series is the distance between that time series and the corresponding shapelet. 

Shapelet transform classification \cite{hills2014STclassification} is a generalization of shapelet decision trees. This algorithm separates the shapelet selection step from the classification step, allowing shapelet approaches to be applied with any supervised classifier instead of being tied to the decision tree model. \cite{hills2014STclassification} also shown that coupling shapelet transformation with a good classifier, particularly an ensemble classifier, significanly improves the classification accuracy. An optimization of shapelet transform classification has been proposed by \cite{bostrom2017binary}. It reduces the execution time of the shapelet selection step, and optimizes the shapelet evaluation for multiclass classification.

The greatest limitation of shapelet-based classification approaches is their time complexity during the training step, which is in the order of $O(n^2m^4)$, where $n$ is the number of time series in the training dataset and $m$ is the length of the longest series. Despite the high time complexity of shapelet algorithms, they are very appreciated because they are easy to interpret, robust to outliers and fast at classifying a new instance. In section \ref{sec:u_shapelet_classification}, we describe the uncertain shapelet transform algorithm, a shapelet approach for the classification of uncertain time series.

\section{UED: a new uncertain dissimilarity measure}\label{sec:UED}
As stated by \cite{err_analysis_jrtaylor}, uncertainty is different from error since it cannot be eliminated, but it can be reduced up to a certain magnitude. Regardless of the measurement method, there is always an uncertainty and uncertain measures cannot be compared with a $100\%$ reliability: the result of the comparison of uncertain values should also be uncertain. 

From now on, we consider only PDF-based represented uncertain values. Let $x$ be an uncertain values, we have $x = \hat{x} \pm \delta{x}$, the exact value of $x$ follows a probability distribution and lies in the interval $[\hat{x} - \delta{x}, \hat{x} + \delta{x}]$. $\hat{x}$ is the best guess of the exact value of $x$. Let $y$ be another uncertain value, any mathematical operator applied on $x$ and $y$ produces a new uncertain value. We have the following uncertainty propagation properties \cite{err_analysis_jrtaylor}:
\begin{itemize}
	\item $z=x+y=\hat{z} \pm \delta{z}$, where $\hat{z}=\hat{x}+\hat{y}$ and $\delta{z}=\delta{x}+\delta{y}$
	\item $z=x-y=\hat{z} \pm \delta{z}$, where $\hat{z}=\hat{x}-\hat{y}$ and $\delta{z}=\delta{x}+\delta{y}$
	\item $z=x^n=\hat{z} \pm \delta{z}$, where $\hat{z}=\hat{x}^n$ et $\delta{z}=|n \frac{\delta{x}}{\hat{x}} \hat{x}^n|$\\
\end{itemize}

Euclidean distance (ED) is widely used in the literature to measure the dissimilarity between time series. It is particularly used in shapelet-based approaches \cite{ye2009time,hills2014STclassification,bagnall2017great}. Using the uncertainty propagation properties, an uncertain dissimilarity measure based on ED can be computed for two uncertain time series $T_1$ and $T_2$ by propagating uncertainty in the ED formula. We name the obtained measure UED for Uncertain Euclidean Distance, and it is defined as follows:
\begin{small}
	\begin{equation}\label{eq:ued} 
	UED(T_1, T_2) = \sum_{i=1}^{n}(\hat{t_{1i}}-\hat{t_{2i}})^2 \pm 2\sum_{i=1}^{n}|\hat{t_{1i}}-\hat{t_{2i}}| \times (\delta{t_{1i}} + \delta{t_{2i}})
	\end{equation}
\end{small}
where $\hat{T_i}$ is the time series of the best guesses of $T_i$.

The ouput of UED is an uncertain measure representing the similarity between the two uncertain time series given as inputs. In order to use UED to classify time series, especially with a shapelet algorithm, an ordering relation for the set of uncertain measures is needed. We propose three ways to compare uncertain measures: the first one is the simpler one and is based on confidence, the second one is a stochastic order and the last one is an interval number ordering.

\subsection*{Simple ordering for uncertain measures}\label{subsec:simple_ordering}
This ordering is based on two simple properties. Let $x$ and $y$ be two PDF-based uncertain measures, the first property is the property of equality and states that two uncertain measures are equal if their best guesses and their uncertainties  are equals.
\begin{equation}
x = y \iff \hat{x} = \hat{y} \land \delta{x} = \delta{y}
\end{equation}

The property of inferiority is the second one and states that the uncertain measure $x$ is smaller than the uncertain measure $y$ if and only if the best guess of $x$ is smaller than the best guess of $y$. In the case where $x$ and $y$ have the same best guesses, the smaller is the one with the smallest uncertainty.
\begin{equation}
x<y \iff(\hat{x} < \hat{y}) \lor ((\hat{x} = \hat{y}) \land (\delta{x} < \delta{y}))
\end{equation}
Unlike the property of equality which is straightforward, the property of inferiority need some explanations. Unfortunately, we don't have a mathematical justification of this property but it is guided by two points: firstly we are in some way confident about the best guess since it must have been given by an expert, and secondly we are more confident with smaller uncertainties.

Of course, these properties do not always give a correct ordering; in fact, if $ x = 2 \pm 0.5 $ and $ y = 2 \pm 0.1 $ then the inferiority property says that $ y < x $. Now, if there is an oracle able to compute the exact value of any uncertain measure, it might says that $x=1.8$ and $y=2$, and thus invalidating our ordering. This observation also holds for the properties of equality.

\subsection*{Stochastic ordering of uncertain measures}\label{subsec:st_ordering}
An uncertain measure can be considered as a random variable with mean equals to the best guess and standard deviation equals to the uncertainty. Given this consideration, a stochastic order can be defined on the set of uncertain measures. A random variable $X$ is \textit{stochastically less than or equal to} (noted $ \leq_{st} $) another random variable $Y$ if and only if $ Pr[X > t] \leq Pr[Y > t] \; \forall t \in \mathbb{I} $, where $ \mathbb{I} $ is the union of the domains of $X$ and $Y$ \cite{Marshall2011}. The stochastic order can be rewritten and developed as follows:
\begin{equation}\label{stochastic_order}
\begin{split}
X \leq_{st} Y & \iff Pr[X > t] \leq Pr[Y > t] \; \forall t \in \mathbb{I} \\
& \iff 1 - Pr[X > t] > 1 - P[Y > t] \; \forall t \in \mathbb{I} \\
& \iff Pr[X \leq t] > Pr[Y \leq t] \; \forall t \in \mathbb{I} \\
& \iff CDF_X(t) > CDF_Y(t) \; \forall t \in \mathbb{I} \\
\end{split}
\end{equation}
$CDF_X(t)$ is the cumulative distribution function of the random variable $X$ evaluated at $t$. Because the size of $\mathbb{I}$ is infinite, we discretized it as being the set of the following values: 
\begin{equation}\label{eq:discretization}
\begin{split}
\min(\mathbb{I}) + i \times \frac{\max(\mathbb{I}) - \min(\mathbb{I})}{k}\\
\text{$0 \leq i \leq k$ and $k$ is a whole number to be defined.} 
\end{split}
\end{equation}

Unlike the simple ordering which is a total order, the stochastic ordering is a partial order. That means, the relation \textit{stochastically less than or equal to} is not defined for any two random variables, and thus any two uncertain measures can not be sorted using this stochastic order. This is clearly a limitation, but we did not find a total stochastic ordering in the literature.

\subsection*{Interval numbers ordering}
\begin{definition}[Interval number] An interval number $i_n$ is a number represented as an interval, that is $i_n=[i_n^l,i_n^u]$, where $i_n^l$ and $i_n^u$ are respectively the lowest and highest possible values of the number. 
\end{definition}

A PDF-based uncertain value is by definition an interval number, enhanced with the distribution of the exact value and a best guess. Given two uncertain values $x$ and $y$, the interval number-based ordering can be estimated using the following probability \cite{Xu2002,Yue2011}:
\begin{equation}\label{eq:interval_ordering}
Pr[x \geq y] = \max(1 - \max(\frac{\hat{y}+\delta y - \hat{x} + \delta x}{2\delta x + 2\delta y}))
\end{equation}

Unlike the stochastic ordering, the simple ordering and the interval number-based ordering do not exploit the uncertainty distribution, nor the best guess given by the expert. The simple ordering required the best guess to be known, and this is not always the case in practice. If the compared uncertain values do not overlap at all, that is they do not have some possible exact values in common, all the three ordering give the same order. 

Now that we know how to sort uncertain measures, let us see how to use UED to classify uncertain time series.

\section{UST: The uncertain shapelet transform classification}\label{sec:u_shapelet_classification}
In this section, we describe how to classify uncertain time series using shapelets. Uncertain observations are represented using the probability density function model (or simply pdf model). We start by defining the concepts that are used in our algorithm, then we describe the algorithm itself.

\subsection{Definition of concepts}

\begin{definition}[Uncertain subsequence] An uncertain subsequence $S$ of an uncertain time series $T$ is a series of $l$ (its length) consecutive uncertain values in $T$. 
	\begin{equation}
	\begin{split}
	S = \hat{S} \pm \delta{S}  = \{t_{i+1} \pm \delta{t_{i+1}}, ..., t_{i+l} \pm \delta{t_{i+l}}\} \\
	1 \leq i \leq m-l, \; 1 \leq l \leq m, m = |T|
	\end{split}
	\end{equation}
\end{definition}

\begin{definition}[Uncertain dissimilarity] The dissimilarity between two uncertain subsequences $S$ and $R$ is the uncertain distance between them
	\begin{equation}
	d = UED(S,R) = UED(R,S)
	\end{equation}
	
	The dissimilarity between an uncertain time series $T$ and an uncertain subsequence $S$ is the dissimilarity between $S$ and the subsequence of $T$ that is the most similar to $S$. It is formally defined as follows:
	\begin{equation}
	UED(T,S)=\min \{ UED(S,R) \;| \; \forall R \subseteq T, |S|=|R|\}
	\end{equation}
\end{definition}

\begin{definition}[Uncertain separator] An uncertain separator $sp$ for a dataset $D$ of uncertain time series is an uncertain subsequence that divides $D$ in two parts $D_1$ and $D_2$ such that:
	\begin{equation}
	\begin{split}
	D_1 = \{ T\; | UED(T,sp) \le \epsilon\:, \forall T \in D \}\\ 
	D_2=\{ T\; | UED(T,sp) > \epsilon\:, \forall T \in D \}
	\end{split} 
	\end{equation}
\end{definition}

As in \cite{hills2014STclassification}, the quality of a separator is measured using the information gain (IG). Given the previous definitions, we can give a formal definition of an uncertain shapelet.
\begin{definition}[Uncertain shapelet] An uncertain shapelet $S$ for a dataset $D$ of uncertain time series is an uncertain separator that maximized the information gain.
	\begin{equation}
	S=argmax_{sp}(IG(D,sp))
	\end{equation}
\end{definition}

\subsection{Uncertain shapelet transform classification}
Our algorithm for uncertain time series classification is an extension of the shapelet transform algorithm \cite{hills2014STclassification}. 

Given a dataset $D$ of uncertain time series, the first step is to select the top $k$ best uncertain shapelets from the dataset. This step is achieved using the procedure described by Algo. \ref{alg:shapelet_selection}, which takes as input, the dataset $D$, the maximum number of uncertain shapelets to be extracted $k$, the minimum and the maximum length of an uncertain shapelet $MIN$ and $MAX$. This algorithm uses three subprocedures:
\begin{itemize}
	\item $GenCand(T,MIN,MAX)$ which generates every possible uncertain shapelet candidates from the input uncertain time series $T$. These candidates are uncertain subsequences of $T$, with length at least $MIN$ and at most $MAX$.
	\item $AssessCand(cands, D)$ which computes the quality of each candidate in the list of candidates $cands$. The quality of a candidate is the information gain it produces when used as a separator for the dataset.Cameroun
	\item $ExtracBest(C, Q, k)$ which takes the list of uncertain shapelet candidates $C$, their associated qualities $Q$ and returns first $k$ uncertain shapelets with highest qualities.
\end{itemize}
In summary, Algo. \ref{alg:shapelet_selection} generates every uncertain subsequences of length at least $MIN$ and at most $MAX$ from the dataset, assesses the quality of each one by computing the information gain obtained when it is used as a separator for the dataset and finally returns the $k$ subsequences that produce the highest information gain. The parameters $MIN$ and $MAX$ should be optimized to reduce the execution time of the algorithm. With the knowledge of the domain, the length of a typical shapelet can be estimated and used to set $MIN$ and $MAX$ in order to reduce the number of shapelet candidates. By default $MIN$ is set to $3$ and $MAX$ is set to $m-1$, where $m$ is the length of the time series.

\begin{algorithm}
	\caption{Top-K Uncertain Shapelet Selection}
	\label{alg:shapelet_selection}
	\begin{algorithmic}[1]
		\Function{UShapeletSelection}{$D,k,MIN, MAX$}
		\State $C \leftarrow \emptyset; Q \leftarrow \emptyset$
		\For{$i \gets 1,n$}
		\State $cands \leftarrow GenCand(T_i,MIN,MAX)$
		\State $qualities \leftarrow AssessCand(cands, D)$
		\State $C \leftarrow C + cands$
		\State $Q \leftarrow Q + qualities$
		\EndFor
		\State $S \leftarrow ExtractBest(C, Q, k)$
		\State \Return S \Comment{Top k uncertain shapelets}
		\EndFunction
	\end{algorithmic}
\end{algorithm}

The next step after the top-$k$ uncertain shapelets selection is the uncertain shapelet transformation. This step is done using Algo. \ref{alg:shapelet_transform}, which takes as input the dataset $D$, the set of the top-$k$ uncertain shapelets $S$ and the number of uncertain shapelets $k$. For each uncertain time series in the dataset, its uncertain feature vector of length $k$ is computed using $UED$. The $i^{th}$ element of the vector is the $UED$ between the uncertain time series and the uncertain shapelet $i$. Because the uncertainties add up during the transformation, the uncertain feature vectors are such that the scale of the best guesses is smaller than the scale of the uncertainties. It is very important to have everything on the same scale. The second for loop of Algo. \ref{alg:shapelet_transform} performs the standard normalization of the transformed dataset. We use $\hat{D_{:j}}$ to represent the list of the best guesses of uncertain dissimilarities between every uncertain time series and the $j^{th}$ uncertain shapelet, and $\delta{D_{:j}}$ is the list of the corresponding uncertainties. The scaled and transformed dataset is finally returned by the algorithm.

\begin{algorithm}
	\caption{Uncertain Shapelet Transformation}
	\label{alg:shapelet_transform}
	\begin{algorithmic}
		\Function{UShapeletTransformation}{$D,S,k$}
		\For{$i \gets 1,n$}
		\State $temp \leftarrow \emptyset$
		\For{$j \gets 1,k$}
		\State $temp_j \leftarrow UED(T_i, S_j)$
		\EndFor
		\State $D_i \leftarrow temp_j$
		\EndFor
		
		\For{$i \gets 1,n$}
		\For{$j \gets 1,k$}
		\State $best \leftarrow \frac{\hat{D_{ij}} - mean(\hat{D_{:j}})}{std(\hat{D_{:j}})}$
		\State $delta \leftarrow \frac{\delta{D_{ij}} - mean(\delta{D_{:j}})}{std(\delta{D_{:j}})}$					
		\State $D_{ij} = best \pm delta$
		\EndFor
		\EndFor
		
		\State \Return D \Comment{The transformed dataset}
		\EndFunction
	\end{algorithmic}
\end{algorithm}

The third and last step is the effective classification. A supervised classifier is trained on the uncertain transformed dataset, such that, given the feature vector of an unseen uncertain time series, it can predict its class label. Since the uncertainty have been propagated, the training process can be aware of uncertainty by taking it as part of the input. More specifically, best guesses are features and uncertainties are features of best guesses, and thus are metafeatures. There exists many supervised classifiers in the literature for the classification of uncertain tabular data \cite{udm2018,Aggarwal2009}. We have decision tree-based methods \cite{tsang2009decision,Qin2009}, SVM-based methods \cite{Bi2005,Yang2007,li2020robust} and Naive bayes-based methods \cite{Qin2011,Qin2010}. Since the transformed data is an uncertain tabular data, uncertain supervised classifiers can be used for the classification step. Furthermore, any supervised classifier can be used as soon as the transformed dataset is formatted in a way that is accepted by that classifier.

If instead of $UED$, one of the existing metrics from the state of the art (DUST, MUNICH, PROUD or FOTS) is used, the classifier would not be able to learn while being aware of uncertainty in the input since the output of these metrics are apparently $100\%$ reliable; most importantly, it would not be possible to take advantage of an uncertain classifier.

\begin{figure*}[ht]
	\includegraphics[width=\linewidth]{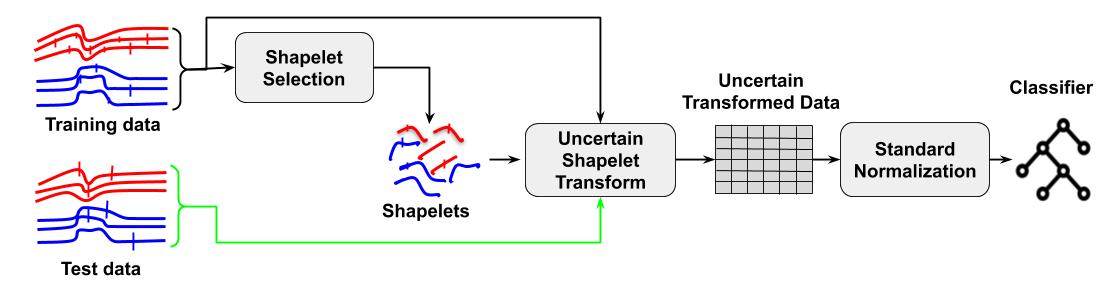}
	\caption{Uncertain time series classification process}
	\label{fig:ust_overview}
\end{figure*}

Fig. \ref{fig:ust_overview} gives an overview of the classification process. During the training step, top-$k$ uncertain shapelets are selected and an uncertain supervised model (illustrated here by a decision tree for simplicity) is trained on the uncertain transformed dataset. During the test step, the uncertain shapelets extracted during the training step are used to transform the test set, and the trained model is used to predict the class labels of the test set according to the result of the transformation. We call this model \textit{UST} for Uncertain Shapelet Transform classification.

Since the number of shapelet candidates of length $l$ in the dataset is bounded to $ n \times (m-l+1) $, the subprocedures used by the Uncertain Shapelet Transform algorithm do not last forever; the rest of the algorithm are for loops, and thus the UST algorithm ends. The algorithm also evaluates all possible shapelet candidates and keep only the $k$ best ones. In term of time complexity, it is the same time complexity as the original shapelet transform (ST) which is $O(n^2m^4)$. In fact the differences are the dissimilarity computation and the transformation process. UST uses UED to compute the dissimilarity, while ST uses the Euclidean distance (ED). The time complexity of UED is $O(n) + O(n)$ which is asymptotically equal to $O(n)$, the time complexity of ED. The scaling of the transformed dataset does not change the complexity order of the transformation process as well. 

\section{Experiments}\label{sec:experiments}
In this section, we experimentally assess our model and compare it with the state of the art. The comparison criterion is the model classification accuracy as it is always done in the literature \cite{bagnall2017great,hills2014STclassification,fots:siyou,uts_sim.1931D}. Since the output of our model is the probability distribution over the set of classes, we take the most probable class as the predicted class and use it to compute the model accuracy. 

\subsection{Compared models}
We have compared different models which are different configurations of the UST model. In particular, our models are built regarding the following attributes:

\subsubsection{uncertain similarity} this is how the dissimilarity between uncertain subsequences is computed by UST. This attribute has five possible values which are ED, UED, FOTS, DUST UNIFORM and DUST NORMAL. 
	
The difference between DUST NORMAL and DUST UNIFORM is the assumption made about the data distribution. Both assume that the uncertainty is normally distributed. DUST UNIFORM assumes that the data in each time series are uniformly distributed. Given this assumption, the distance between two uncertain values $x_i$ and $y_i$ is the following:

\begin{equation}
	dust(x_i, y_i) = \frac{|\hat{x_i}-\hat{y_i}|}{2\sigma}
\end{equation}
	
DUST NORMAL assumes that the data in each time series follow a Gaussian distribution and is defined as follows
\begin{equation}
dust(x_i, y_i) = \frac{|\hat{x_i}-\hat{y_i}|}{2\sigma (1 + \sigma ^2)}
\end{equation} 

In any case, we set $\sigma=\max(\delta x_i, \delta y_i)$. 

We set the parameter of FOTS following the specifications in its original paper \cite{fots:siyou}. Hence, the number of windows ($m$) is equal to the length of a window ($w$) which is equal to half the length of the time series. The time index ($t$) used to compute the autocovariance matrices is equal to half the length of the time series, and finally the number of eigenvectors ($k$) is set to $4$.

\subsubsection{ordering strategy} This is the method used to sort uncertain measures, that is simple, stochastic or interval ordering. When measures are not uncertain (when using FOTS or DUST), we use the natural order.

For the stochastic ordering we consider an uncertain measure $x$ to be normally distributed. Given this assumption, the following cumulative distribution function can be used
\begin{equation}
CDF_X(t) = \frac{1}{2}(1 + erf(\frac{t-\hat{x}}{\delta{}x \sqrt{2}}))
\end{equation}
where $erf(\cdot)$ is the Gauss error function. To discretize $\mathrm{I}$ (using Eq. \ref{eq:discretization}), we fixed the value of $k$ to $100$. Larger values of $k$ lead to best approximation of $\mathrm{I}$, however slow the classification process. We tried several values of $k$, but $k=100$ worked better. We have also used a relaxed version of the stochastic ordering: given two random variables $X$ and $Y$, we have $X \leq_{st} Y$ if the number of values $t$ in $\mathrm{I}$ such that $CDF_X(t) > CDF_Y(t)$ is greater than the number of values $t$ in $\mathrm{I}$ such that $CDF_X(t) \leq CDF_Y(t)$.

For the interval ordering, we say that $x \leq y$ if $Pr[x  \leq y] > 0.5$

\subsubsection{supervised classifier} This is the model used to classify the transformed dataset in the last step of UST. We used the classical Gaussian Naive Bayes (GNB) and the Uncertain Gaussian Naive Bayes (UGNB) models. We implemented UGNB following \cite{Qin2011}. We choose these classifiers for their simplicity, we want to evaluate UED, the importance of propagating uncertainty following by the use of a classifier that takes uncertainty into account during its training phase.

For each model, the parameter $MIN$ and $MAX$ are set to $3$ and $m-1$ respectively, where $m$ is the length of the uncertain time series in the dataset being processed. Because of the high time complexity of the algorithm, we have used a time contract to limit the execution time of each model. After the evaluation of an uncertain shapelet candidate, the next candidate is evaluated only if there is time remaining in the contract; otherwise the shapelet search is ended. Because FOTS is more time consuming than ED, UED and DUST, we set FOTS's time contract $12$ times higher above the time limit of other measures.

Tab. \ref{tab:models} gives a summary of the different models that are evaluated and compared throughout our experiments.
\begin{table*}[ht]
	\centering
	\caption{Summary of the models that are compared in our experiments.}
	\label{tab:models}
	\begin{tabular}{|l|c|c|c|c|}
		\hline
		\textbf{Name} & \textbf{Measure} & \textbf{Ordering} & \textbf{Classifier} & \textbf{Time contract}\\
		\hline
		ST & ED & Natural & GNB & $10$ minutes\\
		\hline 
		UST(DUST\_NORMAL) & DUST NORMAL & Natural & GNB & $10$ minutes \\
		\hline
		UST(DUST\_UNIFORM) & DUST UNIFORM & Natural & GNB & $10$ minutes\\
		\hline
		UST(FOTS) & FOTS & Natural & GNB & $120$ minutes\\
		\hline
		UST(UED, GNB) & UED & Simple, Stochastic and Interval & GNB & $10$ minutes\\
		\hline
		UST(UED, UGNB) & UED & Simple, Stochastic and Interval & UGNB & $10$ minutes\\
		\hline
	\end{tabular}
\end{table*}
In order to apply the model UST(UED, GNB), each uncertain feature vector is flatten such that the first half contains the best guesses and the second half contains the uncertainty deviation. This is required because the Gaussian naive bayes (GNB) classifier does not take uncertainty into account.

\subsection{Datasets}
We used datasets from the well known UCR repository \cite{ucr_repo}. Instead of running our experiment on the whole repository, we use only datasets on which shapelet approaches are known to work well. According to \cite{bagnall2017great}, shapelet approaches are more suitable for electric device, ECG, sensor and simulated datasets. Tab. \ref{tab:datasets} gives a summary of the $15$ shapelet datasets on which we conducted our experiments. The first column is the name of the dataset, the second is the number of instances in the training/test set, the third is the length of time series and the fourth and last column is the number of different classes in the dataset. Each dataset is already split into the training and the test sets on the repository.

\begin{table}[ht]
	\centering
	\caption{Datasets Description}
	\label{tab:datasets}
	\begin{tabular}{|l|c|c|c|c|}
		\hline
		\textbf{Datasets} & \textbf{train/test} & \textbf{length} & \textbf{\#classes}\\
		\hline
		BME & 30/150 & 129 & 3\\
		CBF & 30/900 & 129 & 3\\
		Chinatown & 20/345 & 25 & 2\\
		ECGFiveDays & 23/861 & 137 & 2\\
		ECG200 & 100/100 & 96 & 2\\
		ItalyPowerDemand & 67/1029 & 24 & 2\\
		Plane & 105/105 & 145 & 7 \\
		ShapeletSim & 20/180 & 500 & 2 \\
		SmoothSubspace & 150/150 & 16 & 3\\
		SonyAIBORobotSurface1 & 20/601 & 70 & 2\\
		SonyAIBORobotSurface2 & 27/953 & 65 & 2\\
		SyntheticControl & 300/300 & 60 & 6\\
		Trace & 100/100 & 275 & 5 \\
		TwoLeadECG & 23/1139 & 83 & 2 \\ 
		UMD & 36/144 & 151 & 3\\			       
		\hline
	\end{tabular}
\end{table}

\begin{figure}[htb]
	\centering
	\includegraphics[width=1.1\linewidth]{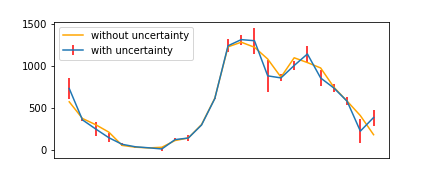}
	\caption{Illustration of uncertainty for an instance from the Chinatown dataset. the uncertainty level is $c=0.6$}
	\label{fig:uncertainty_illustration}
\end{figure}

Since the datasets in this repository are without uncertainty, we manually add uncertainty in the datasets listed in Tab. \ref{tab:datasets}. Given a dataset, the standard deviation $\sigma_i$ of each timestep is computed. For each time series in the dataset, the added uncertainty for the observation at timestep $i$ follows a normal distribution with mean $0$ and standard deviation $ c \times \sigma$, where $\sigma$ is randomly chosen from a normal distribution with mean $0$ and standard deviation $\sigma_i$ . We used different values of $c$ ranging from $0.1$ to $2$. The uncertainty result for an instance from the Chinatown dataset is shown by Fig. \ref{fig:uncertainty_illustration} for $c=0.6$. The orange line is the original time series, and the blue one is the obtained uncertain time series. Sometimes, the original time series does not cross the uncertainty interval (vertical red bars), these cases are there to represent situations where the uncertainty has not been well estimated, maybe because the expert has been too optimistic. Situations where the expert had been too pessimistic are represented by very large uncertainty bars. During the training phase, original time series are not used, only the uncertain time series are used.

We implemented UST in the Python programming language, and we used the open source package sktime \cite{sktime}. The code and the data used for our experiment are publicly available\footnote{\url{https://github.com/frankl1/ustc/releases/tag/litsa}}.

\subsection{Results}
For each of the models we compared, we have recorded the obtained accuracy, the training duration and the testing duration. These values are recorded for each level of uncertainty.

\subsubsection*{Uncertain ordering analysis}
\begin{figure*}[htb]
	\centering
	\begin{subfigure}{0.48\linewidth}
		\includegraphics[width=.98\linewidth]{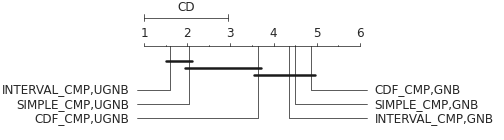}
		\caption{Uncertainty level $c=0.4$}
	\end{subfigure}
	\begin{subfigure}{0.48\linewidth}
		\includegraphics[width=.98\linewidth]{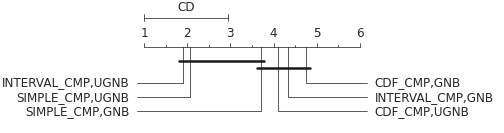}
		\caption{Uncertainty level $c=1.2$}
	\end{subfigure}
	\caption{Critical difference diagrams of UED-based models regarding the ordering strategy for some levels of uncertainty}
	\label{fig:cd_models_acc_per_ordering}
\end{figure*}

Let's analyze how the uncertain ordering used affects the accuracy of UED-based models. Fig. \ref{fig:cd_models_acc_per_ordering} shows the critical difference diagram to exhibit any significant difference between our ordering strategies. The ordering strategy does not affect the model accuracy too much, this is normal given the fact that the three ordering could be different only if the uncertain measures to be sort overlap.  However the simple ordering (called SIMPLE\_CMP on the plots) and the interval ordering (called INTERVAL\_CMP on the plots) are better than the stochastic ordering (called CDF\_CMP on the plots).

\subsubsection*{Accuracy analysis}
In this analysis, UST(UED, GNB) and UST(UED, UGNB) use the interval ordering only. UED-based models are better than the others. They are even better when the uncertain naive bayes is used as classifier, that is UST(UED, UGNB). We observe also that the accuracy of each model decreases when the uncertainty level increases; however the accuracy of UED-based models decreases more slowly than the accuracy of DUST and FOTS-based models: UED-based models are more robust to uncertainty changes, and even more when an uncertain classifier is used. 

\begin{figure*}[htb]
	\centering
	\begin{subfigure}{0.48\linewidth}
		\includegraphics[width=.98\linewidth]{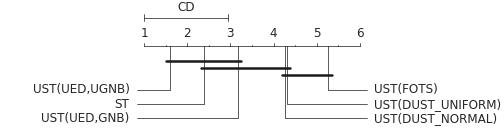}
		\caption{Uncertainty level $c=0.1$}
	\end{subfigure}
	\begin{subfigure}{0.48\linewidth}
		\includegraphics[width=.98\linewidth]{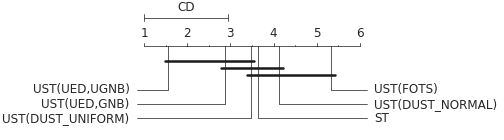}
		\caption{Uncertainty level $c=0.8$}
	\end{subfigure}
	\caption{Critical difference diagrams of models for some levels of uncertainty}
	\label{fig:cd_models_acc}
\end{figure*}

The critical difference diagrams on Fig. \ref{fig:cd_models_acc} show how different the compared models are for some levels of uncertainty. Whatever the level of uncertainty is, propagating the uncertainty gives models that are significantly more accurate than not using uncertainty propagation.

The model ST is the classical shapelet approach, in which the Euclidean distance is used. The classifier used is the Gaussian naive bayes. In this model only the best guesses are used, and there is no uncertainty handling at all. For low uncertainty levels, this model is the second best model, behind UST(UED, UGNB). When the uncertainty increases, ST becomes among the worst models, exhibiting the importance of uncertainty handling in time series classification. For instance, on the dataset SmoothSubspace and the uncertainty level of $c=0.4$, ST achieves $47\%$ of accuracy, UST(UED, GNB) achieves $49\%$, while UST(UED, UGNB) achieves  $83\%$.

\subsubsection*{Training duration}
In term of training duration, the UST model that uses FOTS dissimilarity takes in average $128 \pm 9$ minutes to finish. When using DUST, ED or UED, UST takes in average $10 \pm 0.35$ minutes to finish. Hence, FOTS-based models are much slower compared to DUST and UED-based models. This is because FOTS has a quadratic time complexity (due to eigenvector decomposition) while ED, DUST and UED have a linear time complexity. Hence the FOTS-based model requires more time to learn.

\subsection{Overall discussion}
To build the uncertain shapelet transform classification algorithm, we have handled uncertainty in each step of a typical shapelet algorithm. During the first step where shapelets are selected, UED is used to select shapelets while taking uncertainty into account; when two shapelets has the same information gain, the one that is less uncertain is preferred. During the shapelet transform step, the uncertainty is propagated so that the transformed dataset contains uncertainty information. To really take advantage of the propagated uncertainty, a classifier that has been built for uncertain data is used in the last step. Without the use of an uncertain classifier, all the previous steps of uncertainty management are not really worthy.

Uncertainty is unpredictable, and because we are dealing with it, it is difficult to identify in which uncertain situation our approach will work well. For this reason, we use different levels of uncertainty in our experiment, expecting to cover at most possible situations as we can. The uncertainty levels from $c=1$ to $c=2$ are more likely to be extreme and may not be found in a real application, but it is important to see how the model's behaviour as the uncertainty becomes too large.

We manually added uncertainty in our datasets. Applying our model on a real uncertain dataset will strengthen our contribution. Nevertheless, by using different levels of uncertainty in our datasets, we expect to cover any real situation. 

The uncertain classifier we used is the uncertain naive Bayes. We choose it for its simplicity. There are other uncertain classifiers  \cite{udm2018,Aggarwal2009}, and they can be used in UST, but we did not try them because our goal was to show how important it is to correctly handle uncertainty in the context of uncertain time series classification. We highly recommend to try other uncertain classifiers when in real application.

Finally, the time contract we set during our experiments limits in some ways the discovery of more, and why not better uncertain shapelets. In fact, maybe new uncertain shapelets might have been discovered with a larger time contract.

\section{Conclusion}\label{sec:conclusion}
The goal of this paper was to classify uncertain time series using the shapelet transformation approach. To achieve this goal, we use uncertainty propagation techniques to define an uncertain dissimilarity measure called $UED$. Then we adapt the well known shapelet algorithm to the context of uncertain time series using $UED$ and propose the uncertain shapelet transform algorithm (UST). We have run experiments on state of the art datasets. The results show that propagating uncertainty during the shapelet transformation and then using an uncertain classifier lead to a more accurate model for uncertain time series classification. The idea of uncertainty propagation can be used with any dissimilarity measure, and any uncertain supervised classifier can be used in the classification phase. 

As future works, we firstly intend to run our models without limiting the execution time in order to evaluate every uncertain shapelet candidates. Secondly, we will apply our approach on a real uncertain dataset, and then apply uncertainty propagation techniques using other dissimilarity measures. Finally we will examine uncertainty handling in other time series classification approaches.

\bibliographystyle{plain}
\bibliography{ust-icdm2020}

\end{document}